\documentclass[conference]{IEEEtran}
\IEEEoverridecommandlockouts
\usepackage{cite}
\usepackage{amsmath,amssymb,amsfonts}
\usepackage{algorithmic}
\usepackage{graphicx}
\usepackage{multicol}
\usepackage{tabularx}
\usepackage{steinmetz}
\usepackage{mathtools}
\usepackage{booktabs}
\usepackage{textcomp}
\usepackage{xcolor}
\usepackage{wrapfig}
\usepackage{subfig}

\usepackage{amsmath}

\usepackage[utf8]{inputenc}
\usepackage[linesnumbered,ruled,vlined,english,onelanguage]{algorithm2e}
\usepackage{sidecap}

\usepackage{multirow}
\usepackage[font=small]{caption}\usepackage{tikz}

\def\BibTeX{{\rm B\kern-.05em{\sc i\kern-.025em b}\kern-.08em
    T\kern-.1667em\lower.7ex\hbox{E}\kern-.125emX}}
    
\begin{document}

\title{An Automatic and Efficient BERT Pruning for Edge AI Systems
% with Efficient Evaluation
}

\author{\IEEEauthorblockN{ \textsuperscript\ Shaoyi Huang\textsuperscript{[1]}, Ning Liu\textsuperscript{[2]}, Yueying Liang\textsuperscript{[1]}, Hongwu Peng\textsuperscript{[1]}, \\Hongjia Li\textsuperscript{[2]}, Dongkuan Xu\textsuperscript{[3]}, Mimi Xie\textsuperscript{[4]}, Caiwen Ding\textsuperscript{[1]}
}
\IEEEauthorblockA{
\textsuperscript{[1]}University of Connecticut, 
\textsuperscript{[2]}Northeastern University, \\
\textsuperscript{[3]}The Pennsylvania State University, \textsuperscript{[4]}University of Texas at San Antonio}
}

\maketitle

\begin{abstract}
With the yearning for deep learning democratization, there are increasing demands to implement Transformer-based natural language processing (NLP) models on resource-constrained devices for low-latency and high accuracy.
Existing BERT pruning methods require domain experts to heuristically handcraft hyperparameters to strike a balance among model size, latency, and accuracy. In this work, we propose AE-BERT, an automatic and efficient BERT pruning framework with efficient evaluation to select a "good" sub-network candidate (with high accuracy) given the overall pruning ratio constraints. Our proposed method requires no human experts experience and achieves a better accuracy performance on many NLP tasks.
Our experimental results on General Language Understanding Evaluation (GLUE) benchmark show that AE-BERT outperforms the state-of-the-art (SOTA) hand-crafted pruning methods on BERT$_{\mathrm{BASE}}$. On QNLI and RTE, we obtain 75\% and 42.8\% more overall pruning ratio while achieving higher accuracy. On MRPC, we obtain a 4.6 higher score than the SOTA at the same overall pruning ratio of 0.5. On STS-B, we can achieve a 40\% higher pruning ratio with a very small loss in Spearman correlation compared to SOTA hand-crafted pruning methods. Experimental results also show that after model compression, the inference time of a single BERT$_{\mathrm{BASE}}$ encoder on Xilinx Alveo U200 FPGA board has a 1.83$\times$ speedup compared to Intel(R) Xeon(R) Gold 5218 (2.30GHz) CPU, which shows the reasonableness of deploying the proposed method generated subnets of BERT$_{\mathrm{BASE}}$ model on computation restricted devices.

\end{abstract}


\section{Introduction}
The Transformer-based models have witnessed eye-catching success in various fields, 
especially in natural language processing (NLP), 
such as sentiment classification~\cite{xie2020unsupervised}, question answering~\cite{beltagy2020longformer}, text summarization~\cite{zhang2019hibert}, 
% paraphrase detection, machine reading comprehension, 
and language modeling~\cite{kitaev2020reformer}. As one of the 
% earliest and most classical 
representative Transformer models, BERT~\cite{devlin2019bert}, showed significant improvement on the General Language Understanding Evaluation (GLUE) benchmark, a well-known collection of tasks for analyzing natural language understanding systems~\cite{wang2018glue}.

With the yearning for deep learning democratization~\cite{garvey2018framework}, there are increasing demands to implement BERT on resource-constrained devices with low latency and high accuracy. Several works were proposed  
% (i.e., weight parameters) 
to find a sub-network on BERT$_{\mathrm{BASE}}$ model with desirable pruning ratio and accuracy~\cite{NEURIPS2020_b6af2c97,you2019drawing,prasanna2020bert}. However, these methods require domain experts to heuristically handcraft hyperparameters to strike a balance among model size, latency, and accuracy, and therefore are often time-consuming and fail to achieve a globally optimal solution~\cite{blalock2020state}.

On the other hand, in computer vision, automatic weight pruning has been developed ~\cite{he2018amc,liu2020autocompress} to identify the sub-networks for deep neural networks to achieve similarity as human experts. However, existing works are empirically costly. For each time of sampling sub-networks, it takes several training epochs to recover the accuracy of the sampled sub-networks from estimating the performance of the sub-networks roughly. Nevertheless, the large amount of sampled sub-networks may take more than thousands of training epochs during the sampling and evaluation phase~\cite{he2018amc,liu2020autocompress}. 

In addition, the syntax and semantics information of Transformer in the language/text-domain is more sensitive than computer vision tasks. As a result, it necessities an investigation on the automatic weight pruning on state-of-the-art (SOTA) Transformer-based pre-trained language models.

To this end, we develop an automatic pruning framework on the BERT model for various NLP tasks. The proposed framework has three stages, i.e., sample the sub-networks; evaluate the sampled sub-networks without fine-tuning; fine-tune the winning sub-network from the former stage.

{Our contributions are: (i) We develop AE-BERT, an auto weight pruning framework, to identify a "good" sub-net candidate (with high accuracy) without any expert experience given the overall pruning ratio constraints. (ii) We eliminate the need for thousands of training epochs on sub-networks sampling and evaluation on existing auto weight pruning methods~\cite{he2018amc,liu2020autocompress} while achieving a better accuracy performance. 
Experiments show that AE-BERT outperforms the SOTA hand-crafted pruning methods on BERT. On QNLI and RTE, we obtain 75\% and 42.8\% more overall pruning ratio while achieving higher accuracy. On MRPC, we obtain a 4.6 higher score than the SOTA at the same overall pruning ratio of 0.5. On STS-B, we can achieve a 40\% higher pruning ratio with a very small loss
in Spearman correlation 
compared to SOTA hand-crafted pruning methods~\cite{chen2020lottery}.}

\section{Related Work}

\textbf{Transformer-based Models.} Transformer-based models have great advantages of achieving leading results on major Natural Language Processing (NLP) tasks (i.e., question answering~\cite{van2019does}, machine translation~\cite{luong2015effective}, speech recognition~\cite{dong2018speech}). Recently, ViT~\cite{dosovitskiy2020image} and iGPT~\cite{chen2020generative} have competitive performance on Computer Vision related tasks with the state-of-the-art. The success of the Transformer-based models mainly benefits from the multi-head self-attention mechanism which computes the representation of a sequence by relating different positions of it multiple times in parallel~\cite{vaswani2017attention}. Despite the success of Transformer-based models, their gigantic model size (for example, BERT$_{\mathrm{BASE}}$ has 110M parameters, BERT$_{\mathrm{LARGE}}$ has 340M parameters) results in massive computations and thus high latency which hurdles the deployment of them on space-intensive edge devices. 

\textbf{BERT Pruning.}
To alleviate the conflicts between model performance and model size and thus achieve model inference time speed up and space-saving, weight pruning techniques have been applied to the NLP field. For instance, irregular magnitude weight pruning (IMWP) has been evaluated on BERT, where 30\%-40\% weights with a magnitude close to zero are set to be zero~\cite{gordon2020compressing}. Irregular reweighted proximal pruning (IRPP)~\cite{guo2019reweighted} adopts iteratively reweighted $l_1$ minimization with the proximal algorithm and achieves 59.3\% more overall pruning ratio than irregular magnitude weight pruning without accuracy loss.~\cite{dalvi2020analyzing} investigates the model general redundancy and task-specific redundancy on BERT and XLNet~\cite{NEURIPS2019_dc6a7e65}. The Lottery Ticket Hypothesis (proposed by Frankle et al.~\cite{frankle2018lottery} in computer vision), showing that a sub-network of the randomly-initialized network can replace the original network with the same performance, recently has been investigated on BERT~\cite{NEURIPS2020_b6af2c97,you2019drawing,prasanna2020bert} and achieves high sparsity with slight accuracy degradation. ~\cite{qi2021accommodating, qi2021accelerating,peng2021accelerating,huang2021hmc,chen2021re} explores pruning methods of transformer models to reduce the gigantic model size and increase the inference speed on edge devices.

 \textbf{Knowledge Distillation.} Knowledge distillation (KD) is widely used to compress Transformer-based models while maintaining accuracy. As one of the most popular model compressing methods, the core idea of KD is transferring the knowledge from a large model (teacher) to the compressed model with fewer parameters (student).
 ~\cite{sanh2019distilbert, jiao2020tinybert, wang2020minilm} utilized KD to transfer knowledge from full BERT$_{\mathrm{BASE}}$ to shallow subnets with less number of encoder layers. ~\cite{xu2021rethinking, huang2021sparse} adopted KD to transfer knowledge from full dense BERT$_{\mathrm{BASE}}$ to the sparse subnets with the same number of encoder layers. Layer-wise knowledge distillation was shown to be effective in reducing the risk of overfitting and increasing network accuracy in~\cite{huang2021sparse}.

\section{The Proposed Framework}

\subsection{Problem Formulation}

Consider an $N$-layer model, where the weight and bias of $i^\text{th}$ layer are denoted by ${\mathbf{ W}}_{i}$, ${\mathbf{ b}}_{i}$ respectively. Let us denote the loss function of the $N$-layer model as 
% $f \big( \{{\bf{W}}_{n}\}, \{{\bf{b}}_{n} \} \big)$.
$f \big( \{{\bf{W}}_{i}\}_{i=1}^N, \{{\bf{b}}_{i} \}_{i=1}^N \big)$. For the $N$-layer model, the pruning ratio of the $i^\text{th}$ layer is controlled by ${\mathbf{\beta}}_{i}$ (for dense weight, ${\mathbf{\beta}}_{i}$ = 0). Given a specific overall pruning ratio target, we randomly generate $S$ pruning strategies containing varying pruning ratio strategies: ${\mathbf{B}}_{j}$ = (${\mathbf{\beta}}_{0}^j$, ${\mathbf{\beta}}_{1}^j$, ${\mathbf{\beta}}_{2}^j$, ..., ${\mathbf{\beta}}_{N}^j$), $j = 1, ..., S$. Our goal is to find the optimal pruned sub-network such that the loss function is minimized and the performance of the model achieves the best. Thus our problem is formulated as: 

\vspace{-0.1in}

\begin{equation}
\scalebox{0.85}{
% \begin{aligned}
${\mathbf B^*} = \underset{ 
\{{\mathbf W}_{i}\},
\{{\mathbf b}_{i}\},
\{{{\mathbf B}_j}\}
}
{\operatorname{argmin}}
f \big( \{{{\mathbf W}_i}\}_{i=1}^N, 
\{{{\mathbf b}_i \}_{i=1}^N, 
\{{{\mathbf B}_j}} \}
\big), 1\le j\le S$

\label{lossfunction}
% \end{aligned}
}
\end{equation}
where ${\mathbf{B}}_{j}$ is the $j^\text{th}$ pruning ratios combination or the $j^\text{th}$ candidate and $\mathbf{{B^*}}$ is the optimal one among the $S$ generated strategies. The overall pruning ratio of the model forms the constraints for full model pruning. Under such constraints, we generate $S$ pruning strategies which satisfy the pruning requirements or constraints and form a searching space. To find a "good" candidate (with high accuracy) when the given constraints of the overall pruning ratio, the searching space may be huge. The existing methods in \cite{he2018amc,liu2020autocompress} execute several epochs of fine-tuning each candidate before selecting the best candidate or pruning strategy. As a result, an ample searching space forms a heavy workload in this step. Based on this, we propose an automatic BERT compression method which is fine-tuning free between the steps of pruning and the best candidate selection.

\subsection{AE-BERT}

\begin{figure*}[t]
\centering
\includegraphics[width=.90\textwidth]{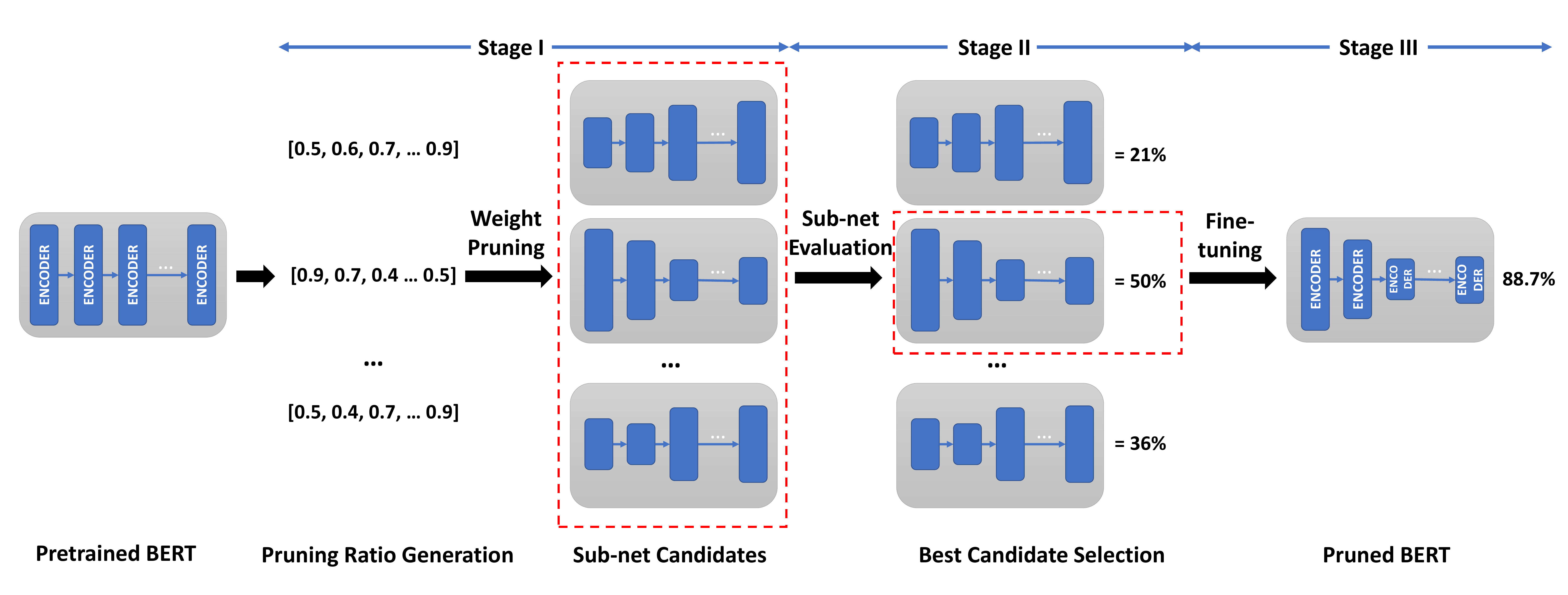}
\vspace{-.1in}
\caption{The three-stage AE-BERT framework.
\vspace{-.2in}
}
\label{fig:workflow}
\end{figure*}

\begin{algorithm}[t]
\footnotesize
% \scriptsize
  \caption{AE-BERT
%   \red{(notation needs to be aligned)}
  }
  \label{alg:AE-BERT}
\begin{algorithmic}
  \STATE {\bfseries Input:} Model $\mathbf{M}$ with a collection of $\mathbf{N}$ weights ($\mathbf{W}_{1}$,$\mathbf{W}_{2}$,...,$\mathbf{W}_{k}$), overall pruning ratio $p$, number of candidates $n$, $best\_accuracy$ = 0, $best\_candidates$ = none
  \STATE {\bfseries Output:} the fine-tuned best candidate $\mathbf{M'}$
  \STATE \textbf{Stage I:} Randomly generate n pruning strategies, $\mathbf{S}_{1}$,$\mathbf{S}_{2}$,...,$\mathbf{S}_{n}$, and each strategy resulting in an overall pruning ratio of $p$
  \FOR{$i=1$ to $n$ }
    \FOR{$j=1$ to $N$ }
        \STATE \textbf{Stage I:} $\mathbf{W}_{j'}$ = magnitude pruning($\mathbf{W}_{j}$)
        
    \ENDFOR
    \STATE $\mathbf{M}_{i}$ = model $\mathbf{M}$ with weights ($\mathbf{W}_{1'}$,$\mathbf{W}_{2'}$,...,$\mathbf{W}_{k'}$)
    \STATE \textbf{Stage II:} $eval\_result$ = evaluate {subnet} $\mathbf{M}_{i}$ 
 
    \IF {$eval\_result \ge best\_accuracy$}
        \STATE $best\_candidate$ = $\mathbf{M}_{i}$
    \ENDIF
  \ENDFOR
  \STATE \textbf{Stage III:} $\mathbf{M'}$ = fine-tuning($best\_candidates$)
\end{algorithmic}
\end{algorithm}

To efficiently identify a good sub-network within the original network, we propose the efficient sub-net evaluation algorithm shown in Figure \ref{fig:workflow} with three stages, i.e., (i) Generate pruning strategies satisfying the constraints and apply the strategies to the original full model. (II) Evaluate the pruned sub-networks without fine-tuning. (III) Fine-tune the highest sub-net evaluation score sub-network.

In stage I, we generate $n$ pruning strategies, and each strategy will result in the same overall pruning ratio of $p$. We adopt the common pruning method, which eliminates the least absolute value of weights~\cite{han2015learning} using the pruning strategies to the original dense model $\mathbf{M}$ and produce $n$ pruned candidates $\mathbf{M}_{1}$, $\mathbf{M}_{2}$, ..., $\mathbf{M}_{n}$, as shown in Algorithm~\ref{alg:AE-BERT}. In stage II, we evaluate the $n$ candidates on the training set, and the one with the highest sub-net evaluation value will be the final $best\_candidate$ after all the iterations. In stage III, we fine-tune the $best\_candidate$ from the former stage and obtain the final "good" sub-network.

For stage II, one mainstream of evaluation is fine-tuning the candidate sub-networks with a small amount of training epochs~\cite{he2018amc,liu2020autocompress,chen2021re,peng2021accelerating}. However, even a small amount of training epoch can be costly since the total number of sampled candidates are extensive because the outer for-loop will amplify the time. To mitigate this cost, we empirically study the correlation between pruning without fine-tuning manner and pruning with fine-tuning manner. We discuss the observation that there exists a high correlation between these two manners in section~\ref{sec:eval}. This high correlation helps efficiently identify the high-quality sub-networks without any training epochs.

\section{Experiments}

\subsection{Datasets}
We adopt GLUE benchmark~\cite{wang2018glue} as our dataset, which consists of three tasks (single sentence tasks, similarity matching tasks, and natural language inference tasks) according to the purpose of tasks and difficulty level of datasets. 
% To be more specific, 
We test our method on two paraphrase similarity matching tasks (MRPC~\cite{dolan2005automatically}, STS-B~\cite{cer-etal-2017-semeval}); and two natural language inference tasks: QNLI~\cite{wang2018glue}, RTE~\cite{wang2018glue}.

\begin{figure*}[!h]
    \centering
    \vspace{-.0in}
    \subfloat[QNLI]{
    \hspace{-.05in}\includegraphics[width=.225\textwidth]{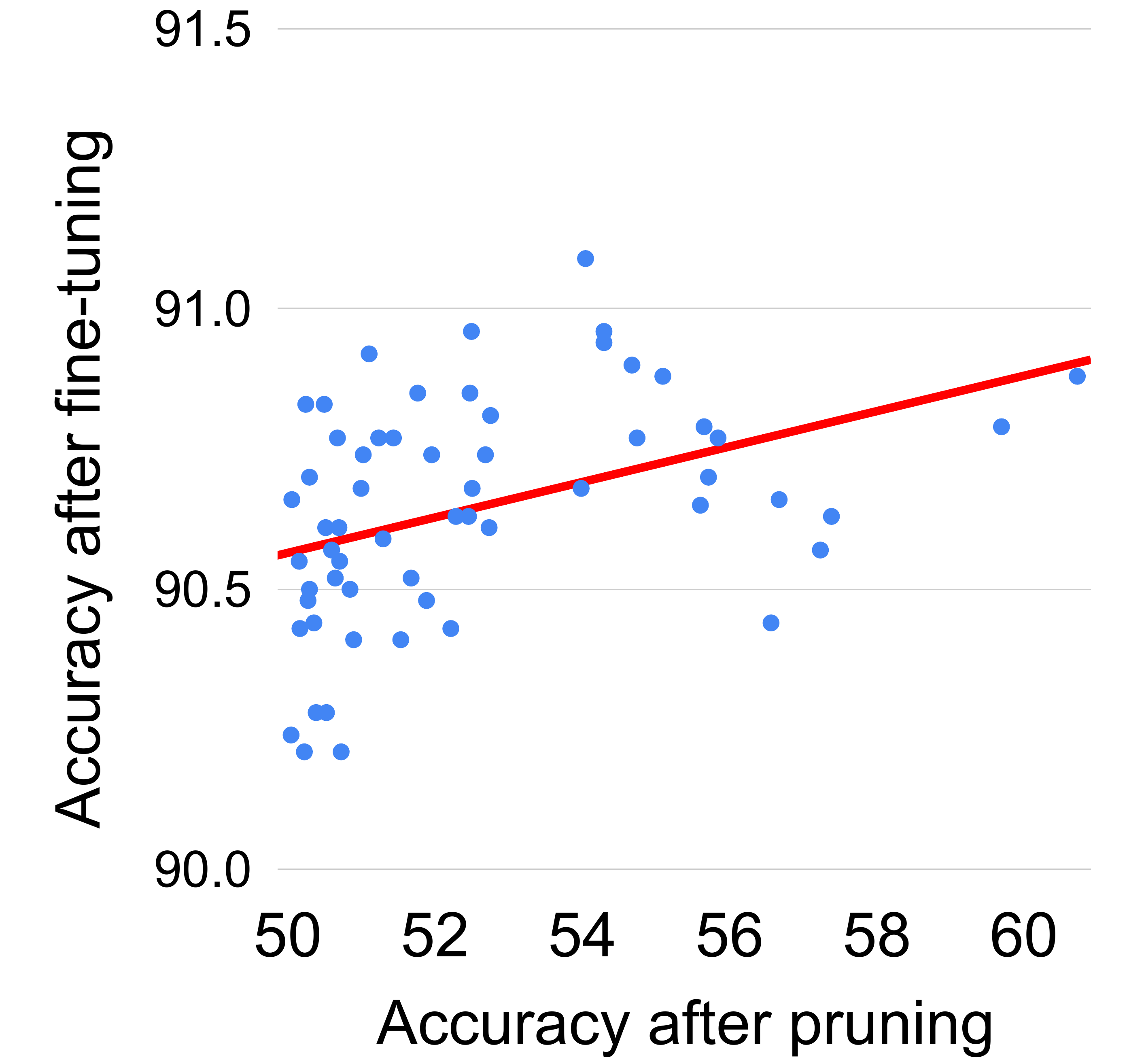}}
    %\caption{fig1}
    \quad
    \subfloat[STS-B]{
    \hspace{-.05in}\includegraphics[width=.225\textwidth]{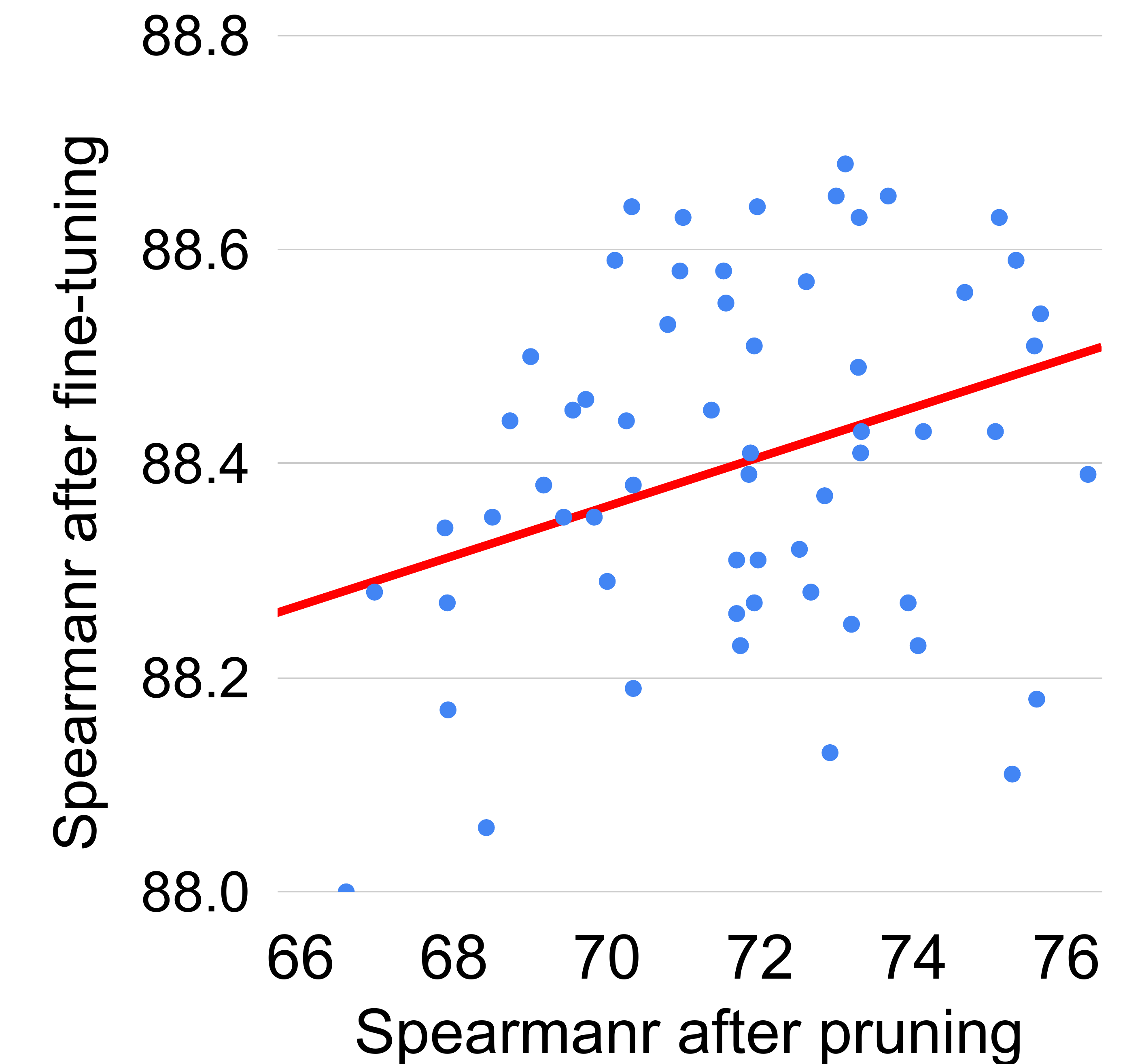}}
    %\caption{fig1}
    \quad
    \subfloat[MRPC]{
    \hspace{-.05in}\includegraphics[width=.225\textwidth]{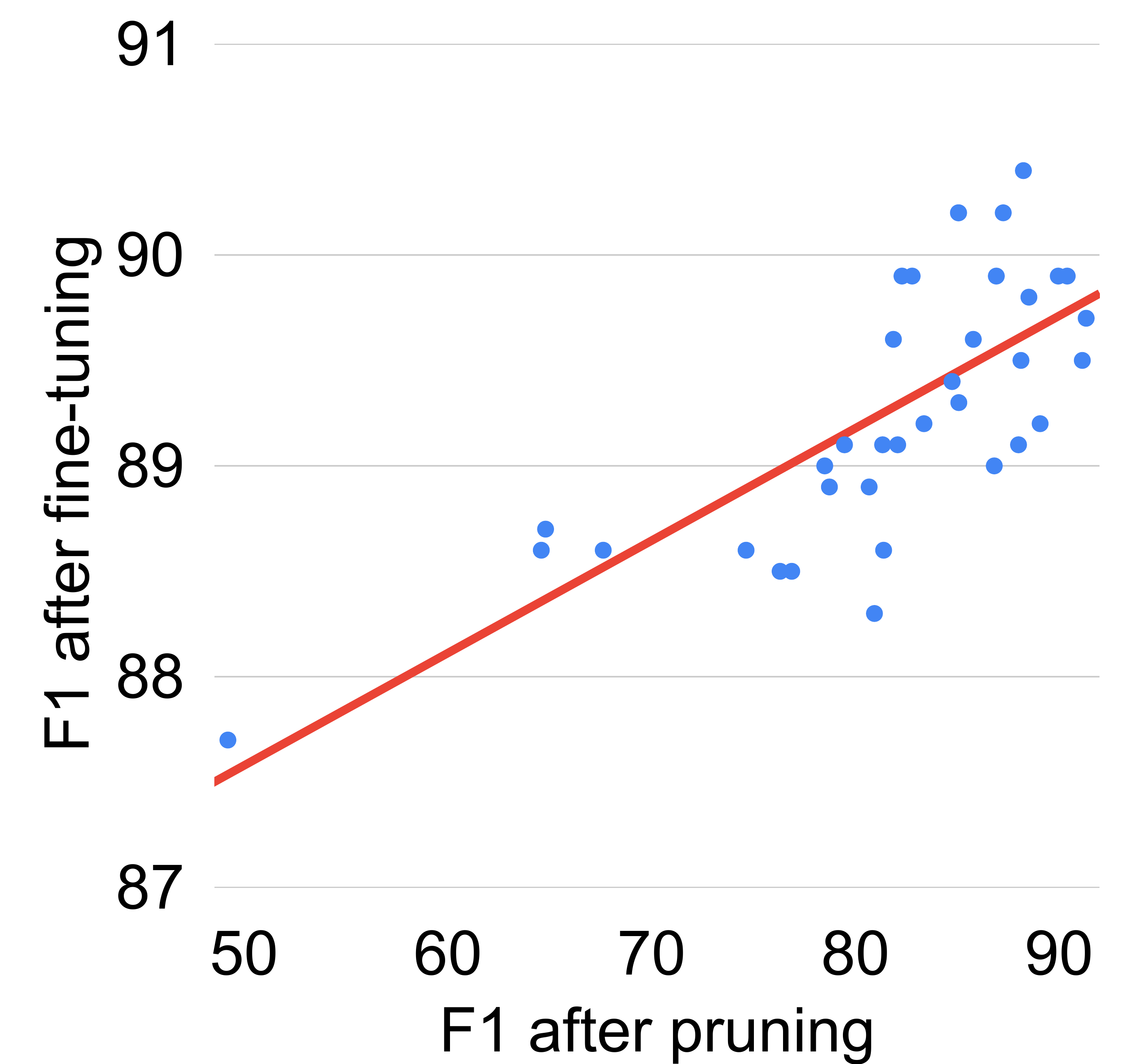}
    %\caption{fig1}
    }
    \quad
    \subfloat[RTE]{
    \hspace{-.05in}\includegraphics[width=.225\textwidth]{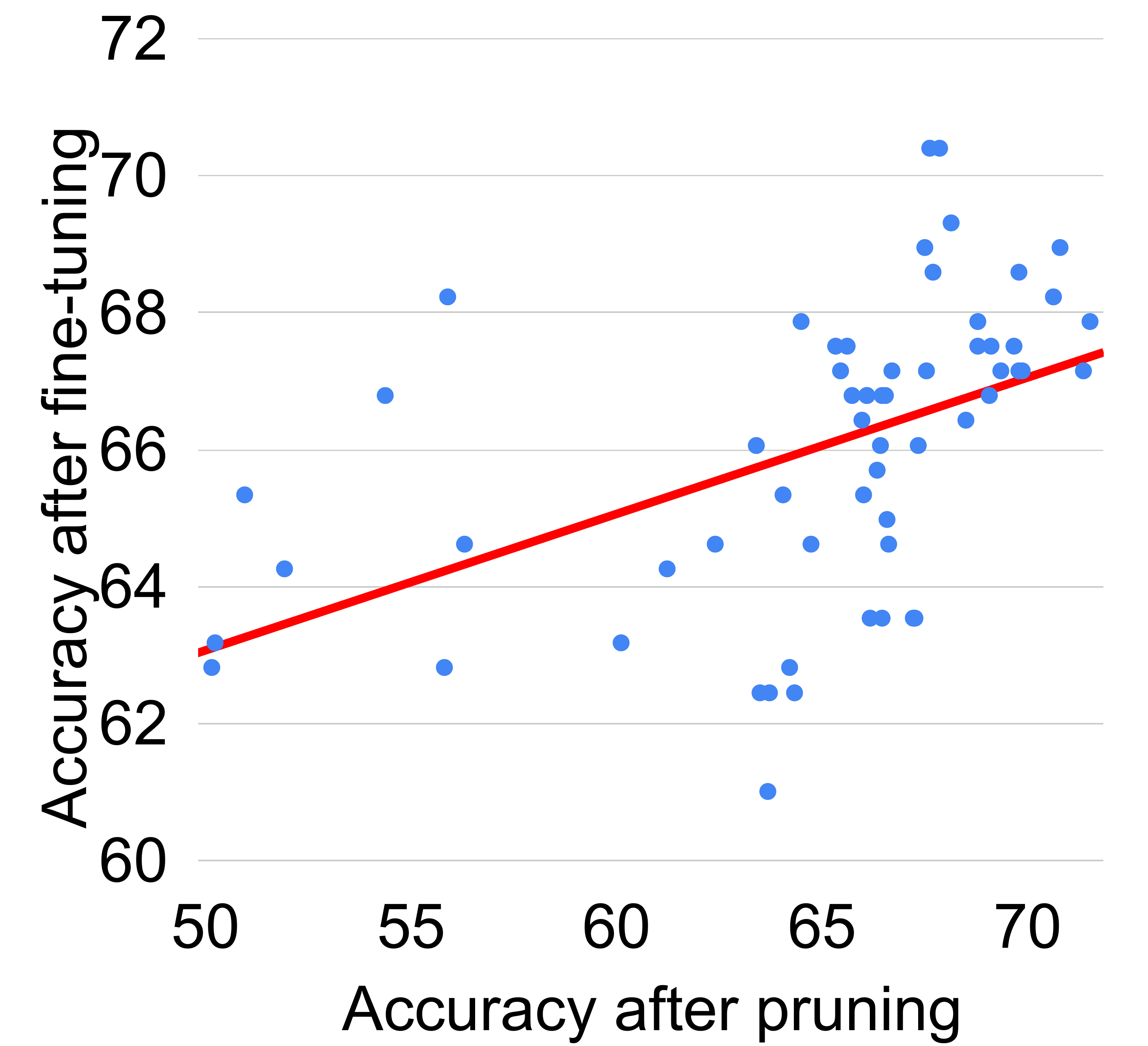}}
    % %\caption{fig1}
    \vspace{-.1in}
    % \caption{Evaluation results on the training set and the dev set of MRPC.}
    % \caption{Comparison of the evaluation results on the training set and the dev set of MRPC.}
    \caption{Correlation of the evaluation results of the pruned and the fine-tuned subnets on four GLUE benchmark tasks. (a). Correlation of after pruning and after fine-tuning accuracy on QNLI; (b). Correlation of after pruning and after fine-tuning Spearman correlation on STS-B; (c). Correlation of after pruning and after fine-tuning F1 score on MRPC; (d) Correlation of after pruning and after fine-tuning accuracy on RTE. The red line in each sub-figure is the trend line.}
    \label{fig:mapping}
    % \vspace{0.1in}
\end{figure*}

\subsection{Experiment Setup}

\textbf{Baseline Models.} The baseline model is our own fine-tuned unpruned BERT$_{\mathrm{BASE}}$~\cite{devlin2019bert}.
We report our results (from the offical bert-base-uncased) 
as Full \textbf{BERT}$_{\mathrm{BASE}}$.
We use the Huggingface Transformer toolkit~\cite{wolf2019huggingface} to conduct our experiment. 
There are 12 layers ($L$ =12; hidden size $H$ = 768; self-attention heads $A$ = 12), with 110 million parameters.
Moreover, we compare our proposed framework with two hand-crafted model compression methods on BERT, {BERT}$_{\mathrm{BASE}}$ irregular~\cite{li2020efficient} and {BERT}$_{\mathrm{BASE}}$ LTH~\cite{chen2020lottery}.
 
\textbf{Metrics.} We apply the same metrics of the tasks as the GLUE paper ~\cite{wang2018glue}, i.e., accuracy scores are reported for RTE and QNLI; F1 scores are reported for MRPC; Spearman correlations are reported for STS-B.
 
\textbf{Platforms.} We use Python 3.6.10 with PyTorch 1.4.0 and CUDA 11.1 on 
Quadro RTX6000 GPU and Intel(R) Xeon(R) Gold 5128 ( 2.30GHz) CPU for software experiments. We use
the Xilinx Alveo U200 board (FPGA) as the hardware platform, which has 4,320 of 18k BRAM, 6,840 DSPs, and 1,882.2k logic cells (LUT) and benefits from the high-level synthesis tool (C/C++). To evaluate the hardware inference performance, we further compare it with Intel i5-5257U (2.7 GHz) CPU in terms of latency and throughput (frame/sequence per second, i.e., FPS).

\textbf{Hyperparameter Selection.} In the fine-tuning stage, we run four epochs for each of the tasks with a batch size of 32 and a learning rate of $3e^{-5}$--$5e^{-5}$ (learning rate that gives the best performance is selected for each task). For the generation of pruning ratio candidates, we randomly generate a list of pruning ratios that give the same overall pruning ratio. For the pruned weight selection, we consider the pruning for encoders. For QNLI, STS-B, MRPC, RTE, we target an overall pruning ratio of 0.7, 0.7, 0.5, and 0.6 as our constraints, respectively.

\textbf{Hardware Setting.} To improve the performance and better utilize the resource during hardware implementation, we adopt operation scheduling methods \cite{wang2018c} for hardware design to allocate an appropriate amount of resources for functions. The optimization problem can be described as: 
\begin{equation}
\small
\begin{aligned}
 \underset{ \{{\mathbf W}_{n}\},\{{\mathbf b}_{n} \}}{\text{\textbf{min}}} &\quad  min(O_1, O_2, ... O_k)
\\ \text{\textbf{subject to}} 
&\quad R_t\geq M \sum_{i=1}^{k} R_i + R_m
\label{optimiz}
\end{aligned}
\end{equation}
where we denote $O_1$, $O_2$, ... $O_k$ as the latency of each individual operation,
$R_t = [R_{DSP}, R_{FF}, R_{LUT}, R_{BRAM}]$ represents the total available resources on an FPGA chip,  $R_i$ stands for the resource that is utilized by each function with an encoder/decoder, and $R_m$ is denoted as the resource used by the DDR controller, PCIe controller, or other types of miscellaneous function within the FPGA system. For the first step, we design hardware without any parallelism. Then we optimize the performance by iteratively adding hardware parallelism in the slowest function or loop and checking the resource constraint until it is satisfied during the process. After optimizing the slowest function or loop, we move to the next slowest function or loop and optimize it with the same procedure until the overall hardware resources and latency are optimal. The hardware scheduling results of each operation (e.g., Matrix Multiplication (MM), dot product attention, add normalization) in encoder and decoder are shown in Table \ref{table:Res_sch}.

\subsection{Results}
\label{sec:eval}
We demonstrate the correlation between pruning without fine-tuning manner and pruning with fine-tuning manner on selected GLUE benchmark tasks in Figure~\ref{fig:mapping}. Each blue dot in all sub-figures represents a candidate, while the values on the x-axis and y-axis stand for the accuracy/metrics values after pruning (pruning without fine-tuning) and the value after fine-tuning (pruning with fine-tuning), respectively. The red line in each subplot is the auto-fitting trend line. The positive slope of trend lines in four sub-figures indicate a linear mapping relation between the manner of pruning without fine-tuning and the manner of pruning with fine-tuning, which justifies our observation.

\begin{table}[!h]
\footnotesize
\centering
	\caption{Comparison of AE-BERT with the state-of-the-arts.}\label{table:main}
\resizebox{1\columnwidth}{!}{
\begin{tabular}{l|l|l|l|l}
\hline
Models & QNLI & STS-B & MRPC & RTE \\ \hline

\begin{tabular}[c]{@{}l@{}}\textbf{Full BERT}$_{\mathrm{BASE}}$ 

\end{tabular}
&91.4 &89.1 &89.4 &71.5  \\ \hline

\begin{tabular}[c]{@{}l@{}}
% \textbf{BERT}$_{\mathrm{BASE}}$
\textbf{BERT$_{\mathrm{BASE}}$  irregular\cite{li2020efficient}} \end{tabular}   
&87.8 &86.7 & - &63.5 \\ 
\textbf{Pruning ratio}   
&0.4 &0.6 &- &0.42 \\ \hline

\begin{tabular}[c]{@{}l@{}}
% \textbf{BERT}$_{\mathrm{BASE}}$
\textbf{BERT$_{\mathrm{BASE}}$  LTH\cite{chen2020lottery}} \end{tabular}   
&88.9 &88.2 & 84.9 &66 \\ 
\textbf{Pruning ratio}   
&0.7 &0.5 &0.5 &0.6 \\ \hline

\begin{tabular}[c]{@{}l@{}}
% \textbf{BERT}$_{\mathrm{BASE}}$ 
\textbf{AE-BERT (ours-Top 1)} \end{tabular}   
&88.7 &86.1 & 89.5 &69.7 \\ 
\textbf{Pruning ratio}   
&0.7 &0.7 &0.5 &0.6 \\ \hline

\begin{tabular}[c]{@{}l@{}}
% \textbf{BERT}$_{\mathrm{BASE}}$ 
\textbf{AE-BERT (ours-Top 10)} \end{tabular}   
& 89.7 &86.7 &89.6 &70.4\\
\textbf{Pruning ratio}   
&0.7 &0.7 &0.5 &0.6 \\ \hline

\hline
\end{tabular}
}
\end{table}

The experimental results of are shown in Table \ref{table:main} on GLUE benchmark tasks. We can achieve the aimed prune ratio with limited accuracy loss by applying our proposed AE-BERT framework. To be more specific, for the paraphrase similarity matching task on MRPC both the top 1 and top 10 F1 scores (89.5 and 89.6) of the sparse model at a pruning ratio of 0.5 even exceed our own fine-tuned baseline (89.4). Furthermore, for STS-B, the Spearman correlation loss for the top 1 result is 3.0, while the top 10 is 2.4 at a pruning ratio of 0.7. On the natural language inference tasks (QNLI and RTE), our proposed framework can limit the accuracy loss up to 1.7 for the top 10 results.

Comparison results with BERT compression works \cite{li2020efficient} and \cite{chen2020lottery} are shown in Table \ref{table:main}, which indicating that our method has advantage on these tasks. (Since there is no result of irregular pruning on MRPC in \cite{li2020efficient}, we do not compare the result of it.) 
Our AE-BERT framework exhibits the highest accuracy performance among various NLP tasks without the need for domain experts' efforts. Under a more overall pruning ratio, our proposed framework achieves higher accuracy than the hand-crafted model compression methods for BERT. On QNLI and RTE, 75\% and 42.8\% more overall pruning ratio is obtained while higher accuracy is maintained. On MRPC, we obtain a 4.6 higher score at the same overall pruning ratio of 0.5. On STS-B, we can achieve a 40\% higher pruning ratio with only up to 2.38\% accuracy loss in Spearman correlation.

A comparison of AE-BERT and sparse BERT$_{\mathrm{BASE}}$ at higher sparsity (0.9) is shown in Table~\ref{table:sparsity_90}. The difference between the two compression methods is that the sparsity of different encoders of the former may be different from each other, while the latter is the same. The experimental results show that AE-BERT can find better subnets than irregular pruned BERT$_{\mathrm{BASE}}$ on QNLI, STS-B, MRPC, and RTE. We also investigate the effects of knowledge distillation on the finetuning stage where we use full BERT$_{\mathrm{BASE}}$ and the best candidates as teacher and student, respectively. We use layer-wise distillation loss instead of hard logits loss to update weights. The experimental results show that KD could further improve the performance of the best candidates. On QNLI, STS-B, MRPC, there are 3\%, 5.8\%, 1.9\% performance gains using KD loss, respectively.

\begin{table}[!h]
\footnotesize
\centering
	\caption{Comparison of AE-BERT (the sparsity of different encoders may vary from each other) with sparse BERT$_{\mathrm{BASE}}$ (the sparsity of different encoders are the same) at same overall sparsity 0.9.}\label{table:sparsity_90}
\resizebox{1\columnwidth}{!}{
\begin{tabular}{l|l|l|l|l}
\hline
Models & QNLI & STS-B & MRPC & RTE \\ \hline

\begin{tabular}[c]{@{}l@{}}\textbf{Full BERT}$_{\mathrm{BASE}}$ 
% \textbf{(ours)
% } 
\end{tabular}
&91.4 &89.1 &89.4 &71.5  \\ \hline

\begin{tabular}[c]{@{}l@{}}
% \textbf{BERT}$_{\mathrm{BASE}}$
\textbf{Sparse BERT$_{\mathrm{BASE}}$} \end{tabular}   
&64.2 &29.5 &79.1 &55.6 \\ 
% \textbf{Pruning ratio}   
% &0.9 &0.9 &0.9 &0.9 \\ 
\hline

\begin{tabular}[c]{@{}l@{}}
% \textbf{BERT}$_{\mathrm{BASE}}$ 
\textbf{AE-BERT} \end{tabular}   
&72.3 &41.2 &80.8 &56.8 \\ 
% \textbf{Pruning ratio}   
% &0.9 &0.9 &0.9 &0.9 \\ 
\hline

\begin{tabular}[c]{@{}l@{}}
% \textbf{BERT}$_{\mathrm{BASE}}$ 
\textbf{AE-BERT + KD} \end{tabular}   
&74.5 &43.6 &82.3 &57.1 \\ 
% \textbf{Pruning ratio}   
% &0.9 &0.9 &0.9 &0.9 \\ 
\hline

\hline
\end{tabular}
}
\end{table}

\textbf{Hardware Evaluation and Analysis.} DSP, FF, and LUT are the three most important metrics while evaluating performance on hardware. In our experiments, we observe from the table that a single encoder occupies 2050, 558.4k, and 692.8k resources for DSP, FF, and LUT, respectively, as shown in Table~\ref{table:Res_sch}. As for the total hardware resources utilization, DSP, FF, and LUT require up to 6840, 2364.5k, and 1882.2k, and the percentages of the resource utilization on FPGA are 30.1\%, 23.6\% and 36.8\%. The latency of an encoder is reduced to 17.23 ms after optimization, which satisfies the real-time constraints, thus making it possible for various NLP tasks while operating on resource-constrained devices.

\begin{table}[!h]   
    \centering
    \caption{Encoder implementation (sparsity = 0.9, batch size = 1)}
     \label{table:Res_sch}
    \resizebox{1\columnwidth}{!}{\begin{tabular}{rrrrr}
    \hline
    \multicolumn{1}{|c|}{\textbf{ }} & \multicolumn{1}{c|}{DSP} & \multicolumn{1}{c|}{FF} & \multicolumn{1}{c|}{LUT} & \multicolumn{1}{c|}{Latency}\\\hline \hline
    \multicolumn{1}{|c|}{\textbf{Total hardware resources}} & \multicolumn{1}{c|}{6,840} & \multicolumn{1}{c|}{2,364.5k} & \multicolumn{1}{c|}{1,882.2k} & \multicolumn{1}{c|}{N/A}\\\hline \hline
    \multicolumn{1}{|c|}{\textbf{Encoder}} & \multicolumn{1}{c|}{DSP} & \multicolumn{1}{c|}{FF} & \multicolumn{1}{c|}{LUT} & \multicolumn{1}{c|}{Latency}\\ \hline
    \multicolumn{1}{|c|}{Sparse MM accelerator 1} & \multicolumn{1}{c|}{662} & \multicolumn{1}{c|}{300.8k} & \multicolumn{1}{c|}{301.6k} & \multicolumn{1}{c|}{5.760 ms}\\\hline
    \multicolumn{1}{|c|}{Dot product attention $\times$ 12} & \multicolumn{1}{c|}{584} & \multicolumn{1}{c|}{119.2k} & \multicolumn{1}{c|}{203.4k} & \multicolumn{1}{c|}{2.770 ms}\\\hline
    \multicolumn{1}{|c|}{Sparse MM accelerator 2} & \multicolumn{1}{c|}{336} & \multicolumn{1}{c|}{46.6k} & \multicolumn{1}{c|}{62.4k} & \multicolumn{1}{c|}{3.520 ms}\\\hline
    \multicolumn{1}{|c|}{Add normalization 1} & \multicolumn{1}{c|}{124} & \multicolumn{1}{c|}{35.8k} & \multicolumn{1}{c|}{36.6k} & \multicolumn{1}{c|}{1.605 ms}\\\hline
    \multicolumn{1}{|c|}{Sparse MM accelerator 3} & \multicolumn{1}{c|}{344} & \multicolumn{1}{c|}{56k} & \multicolumn{1}{c|}{52.2k} & \multicolumn{1}{c|}{1.965 ms}\\\hline
    \multicolumn{1}{|c|}{Add normalization 2} & \multicolumn{1}{c|}{124} & \multicolumn{1}{c|}{35.8k} & \multicolumn{1}{c|}{36.6k} & \multicolumn{1}{c|}{1.605 ms}\\\hline
    \multicolumn{1}{|c|}{\textbf{Resources for 1 encoder}} & \multicolumn{1}{c|}{2050} & \multicolumn{1}{c|}{558.4k} & \multicolumn{1}{c|}{692.8k} & \multicolumn{1}{c|}{17.23 ms}\\\hline 
    \multicolumn{1}{|c|}{\textbf{Percentage}} & \multicolumn{1}{c|}{30.1\%} & \multicolumn{1}{c|}{23.6\%} & \multicolumn{1}{c|}{36.8\%} & \multicolumn{1}{c|}{N/A}\\\hline 
    \hline
    \vspace{-0.1in}
    \end{tabular}}
\end{table}

\textbf{Cross Platform Comparison.}
As shown in Table~\ref{table:platform}, we compare the latency and throughput of a single encoder with a batch size of 1 on Intel(R) Xeon(R) Gold 5218 (2.30GHz) CPU and Xilinx Alveo U200 FPGA board, respectively. The latency/throughput of a single encoder for the CPU and the FPGA are 31.50/31.75, 17.23/58.04, individually. As a result, a 1.83 $\times$ inference speedup on a single encoder compared to the CPU. In conclusion, the proposed pruning method can generate well-performed subnets, and the hardware design on FPGA accelerates the inference time of BERT$_{\mathrm{BASE}}$.

\begin{table}[!h]
\footnotesize
\centering
	\caption{Comparison of inference performance of a single encoder on different hardware devices}\label{table:platform}
\resizebox{1\columnwidth}{!}{
\begin{tabular}{l|l|l}
\hline
Hardware & Latency (ms) & Throughput (FPS)\\ \hline

\begin{tabular}[c]{@{}l@{}}\textbf{Intel(R) Xeon(R) Gold 5218 (2.30GHz) CPU}
% \textbf{(ours)
% } 
\end{tabular}
& 31.50 & 31.75  \\ \hline

\begin{tabular}[c]{@{}l@{}}\textbf{Xilinx Alveo U200 FPGA board}
% \textbf{(ours)
% } 
\end{tabular}
& 17.23 & 58.04  \\ \hline

\hline
\end{tabular}
}
\end{table}

\section{Conclusion}

In this work, we propose a three-stage AE-BERT method to achieve automatic pruning on BERT$_{\mathrm{BASE}}$ without the experience of human experts. In stage I, we sample the sub-networks according to the strategy generated under pruning constraints. In stage II, we evaluate the sampled sub-networks and select the candidate with the best performance on GLUE. Then, we consider this sub-network with the highest potential to achieve the highest performance.
In addition, we achieve fine-tuning free in this stage. We fine-tune the winning candidate from the previous stage and obtain the final sparse model in stage III. Experiments on four GLUE benchmark tasks show that our proposed method outperforms the SOTA hand-crafted pruning methods. Experiments on hardware platforms show that there is a 1.83$\times$ inference time speedup of a single encoder on FPGA compared to CPU.

{\vspace{\baselineskip}
\bibliographystyle{IEEEtran}
\bibliography{bibligraphy}
}

\end{document}